\documentclass{article} 
\usepackage{tencent_ailab_tech_report}
\usepackage[colorlinks = true,
            linkcolor = blue,
            urlcolor  = blue,
            citecolor = blue,
            anchorcolor = blue]{hyperref}   
\usepackage{microtype}
\usepackage{hyperref}
\usepackage{url}
\usepackage{booktabs}
\usepackage{enumitem}
\usepackage{multicol}
\usepackage{multirow}
\usepackage{CJKutf8}
\usepackage{amsmath}
\usepackage{siunitx}
\usepackage{floatflt}
\usepackage{wrapfig}
\usepackage{graphicx}
\usepackage{booktabs}
\usepackage{wrapfig}
\usepackage{authblk}
\usepackage{lipsum}

\usepackage{algorithm}
\usepackage{algorithmicx}
\usepackage{algpseudocode}
\usepackage{microtype}
\usepackage{graphicx}
\usepackage{subfigure}
\usepackage{multirow}
\usepackage{booktabs} 
\usepackage{pifont}  
\usepackage{graphicx}  
\usepackage{subcaption} 
\usepackage{hyperref}

\usepackage{amssymb} 

\algnewcommand{\LeftComment}[1]{\Statex \(\triangleright\) #1}

\usepackage{array}
\usepackage{amsmath}
\usepackage{amssymb}
\usepackage{mathtools}
\usepackage{amsthm}
\usepackage{arydshln}
\usepackage[capitalize,noabbrev]{cleveref}
\usepackage{adjustbox} 
\usepackage{enumitem}

\theoremstyle{plain}

\theoremstyle{definition}

\theoremstyle{remark}

\usepackage[textsize=tiny]{todonotes}

\definecolor{nred}{RGB}{196, 38, 11}
\definecolor{ngreen}{RGB}{18, 141, 21}
\definecolor{nblue}{RGB}{41, 52, 190}
\definecolor{hzw}{RGB}{223, 97, 76}
\definecolor{lt}{RGB}{54, 89, 170}

\newcommand{\ignore}[1]{}

\colmfinalcopy

\title{
Thoughts Are All Over the Place: On the Underthinking of o1-Like LLMs
}

\author[ ]{Yue Wang\thanks{Equal Contribution. The work was done when Yue, Xingyu and Zhiwei were interning at Tencent AI Lab.}~~$^{,1,2}$}
\author[ ]{Qiuzhi Liu$^{*,1}$}
\author[ ]{Jiahao Xu$^{*,1}$}
\author[ ]{Tian Liang$^{*,1}$}
\author[ ]{Xingyu Chen$^{*,1,3}$}
\author[ ]{Zhiwei He$^{*,1,3}$}
\author[ ]{\mbox{Linfeng Song}$^{1}$}
\author[ ]{Dian Yu$^{1}$}
\author[ ]{Juntao Li$^2$}
\author[ ]{Zhuosheng Zhang$^3$}
\author[ ]{Rui Wang$^3$}
\author[ ]{\\ \mbox{Zhaopeng Tu}\thanks{Correspondence to: Zhaopeng Tu \textless zptu@tencent.com\textgreater.}~~$^{1}$}
\author[ ]{\mbox{Haitao Mi}$^{1}$}
\author[ ]{Dong Yu$^{1}$}

\affil[1]{Tencent AI Lab}
\affil[2]{Soochow University}
\affil[3]{Shanghai Jiao Tong University}

\begin{document}

\maketitle

\begin{figure}[h!]
\centering
\vspace{-10mm}
\includegraphics[width=0.8\linewidth]{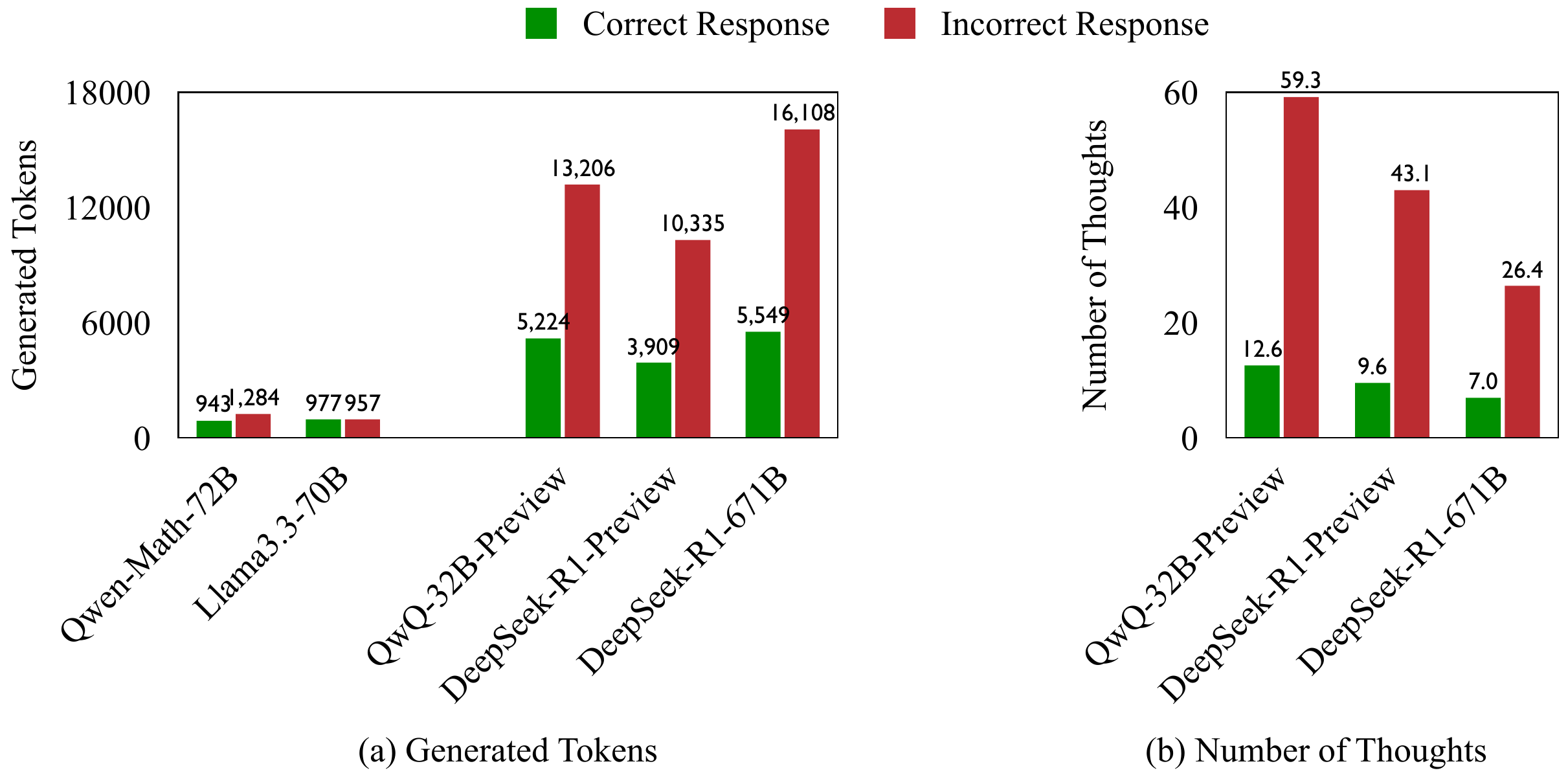}
\caption{Illustration of the {\bf underthinking issue} on the challenging AIME2024 testset: In o1-like models (e.g., QwQ-32B-Preview and DeepSeek-R1-671B), incorrect answers often switch reasoning strategies more frequently than correct ones (Figure b), leading to longer responses without improved accuracy (Figure a). In contrast, conventional LLMs (e.g., Qwen-Math-72B and Llama3.3-70B) show no significant difference in response length between incorrect and correct answers.}
\label{fig:underthinking}
\end{figure} 

\begin{abstract} %
Large language models (LLMs) such as OpenAI's o1 have demonstrated remarkable abilities in complex reasoning tasks by scaling test-time compute and exhibiting human-like deep thinking. However, we identify a phenomenon we term {\bf underthinking}, where o1-like LLMs frequently switch between different reasoning thoughts without sufficiently exploring promising paths to reach a correct solution. This behavior leads to inadequate depth of reasoning and decreased performance, particularly on challenging mathematical problems. 
To systematically analyze this issue, we conduct experiments on three challenging test sets and two representative open-source o1-like models, revealing that frequent thought switching correlates with incorrect responses. We introduce a novel metric to quantify underthinking by measuring token efficiency in incorrect answers. To address underthinking, we propose a decoding strategy with {\color{ngreen} t}hought sw{\color{ngreen} i}tching {\color{ngreen} p}enalty (\textsc{Tip})  that discourages premature transitions between thoughts, encouraging deeper exploration of each reasoning path. 
Experimental results demonstrate that our approach improves accuracy across challenging datasets without requiring model fine-tuning. 
Our findings contribute to understanding reasoning inefficiencies in o1-like LLMs and offer a practical solution to enhance their problem-solving capabilities.
\end{abstract}

\section{Introduction}

Large Language Models (LLMs), such as OpenAI's o1~\citep{openai-learning-to-reason}, have revolutionized artificial intelligence by enabling models to tackle increasingly complex tasks. The o1 model and its replicas~\citep{qwq-32b-preview,DeepSeekAI2025DeepSeekR1IR,MoonshotAI}, known for their deep reasoning capabilities, exemplify the potential of LLMs to exhibit human-like deep thinking by scaling test-time computation during problem-solving. These models aim to explore diverse reasoning strategies, reflect on their decisions, and iteratively refine solutions, closely mimicking human cognitive processes.

Despite their successes, a critical yet underexplored question remains: {\bf Are o1-like LLMs thinking deeply enough?} This study provides an initial exploration of this problem. In this work, we investigate a phenomenon we term {\bf underthinking}, which refers to the tendency of o1-like LLMs to prematurely abandon promising lines of reasoning, leading to inadequate depth of thought. To systematically analyze underthinking, we conduct experiments on three challenging test sets (e.g., MATH500, GPQA Diamond, and AIME2024) and two open-source o1-like models with visible long chains of thought (e.g., QwQ-32B-Preview and DeepSeek-R1-671B). 
Through extensive analyses, we found that underthinking manifests in the following patterns: (1) it occurs more frequently on harder problems, (2) it leads to frequent switching between different thoughts without reaching a conclusion in each, and (3) it correlates with incorrect responses due to insufficient exploration of reasoning paths. For example, Figure~\ref{fig:underthinking} compares the token usage and number of thoughts of correct and incorrect responses. On average, o1-like LLMs consume 225\% more tokens in incorrect responses than in correct ones due to 418\% more frequent thought-switching behaviors.

To quantify this phenomenon, we introduce a novel {\em underthinking metric} that measures token efficiency in incorrect responses by evaluating the proportion of the response that contributes to reaching correct thoughts. Combining the widely-used accuracy metric with the proposed underthinking metric provides a more comprehensive assessment of o1-like models: accuracy measures how often the model can produce {\em correct responses}, while the underthinking metric evaluates the token efficiency within {\em incorrect responses} that contributes to reaching correct thoughts.

In response to these findings, we propose a decoding strategy with {\color{ngreen} t}hought sw{\color{ngreen} i}tching {\color{ngreen} p}enalty (\textsc{Tip}) that discourages premature transitions between thoughts during the generation process. By adjusting decoding penalties for tokens associated with thought switching, the model is encouraged to thoroughly develop each line of reasoning before considering alternatives. Experimental results show that employing \textsc{Tip} improves accuracy across challenging test sets without requiring additional model fine-tuning.

Our study makes the following contributions:
\begin{enumerate}[leftmargin=10pt]
    \item We formally define and characterize the underthinking issue in o1-like LLMs, where models frequently abandon promising reasoning paths prematurely, leading to inadequate depth of reasoning on challenging problems.

    \item We introduce a novel metric to evaluate underthinking by measuring token efficiency in incorrect responses, providing a quantitative framework to assess reasoning inefficiencies.

    \item We propose a decoding approach with thought switching penalty (\textsc{Tip}) that encourages models to deeply explore each reasoning thought before switching, improving accuracy without additional model fine-tuning.
\end{enumerate}

\section{Observing Underthinking Issues}
\label{sec:underthinking}

\begin{figure*}[t!]
    \centering
    \includegraphics[width=0.95\linewidth]{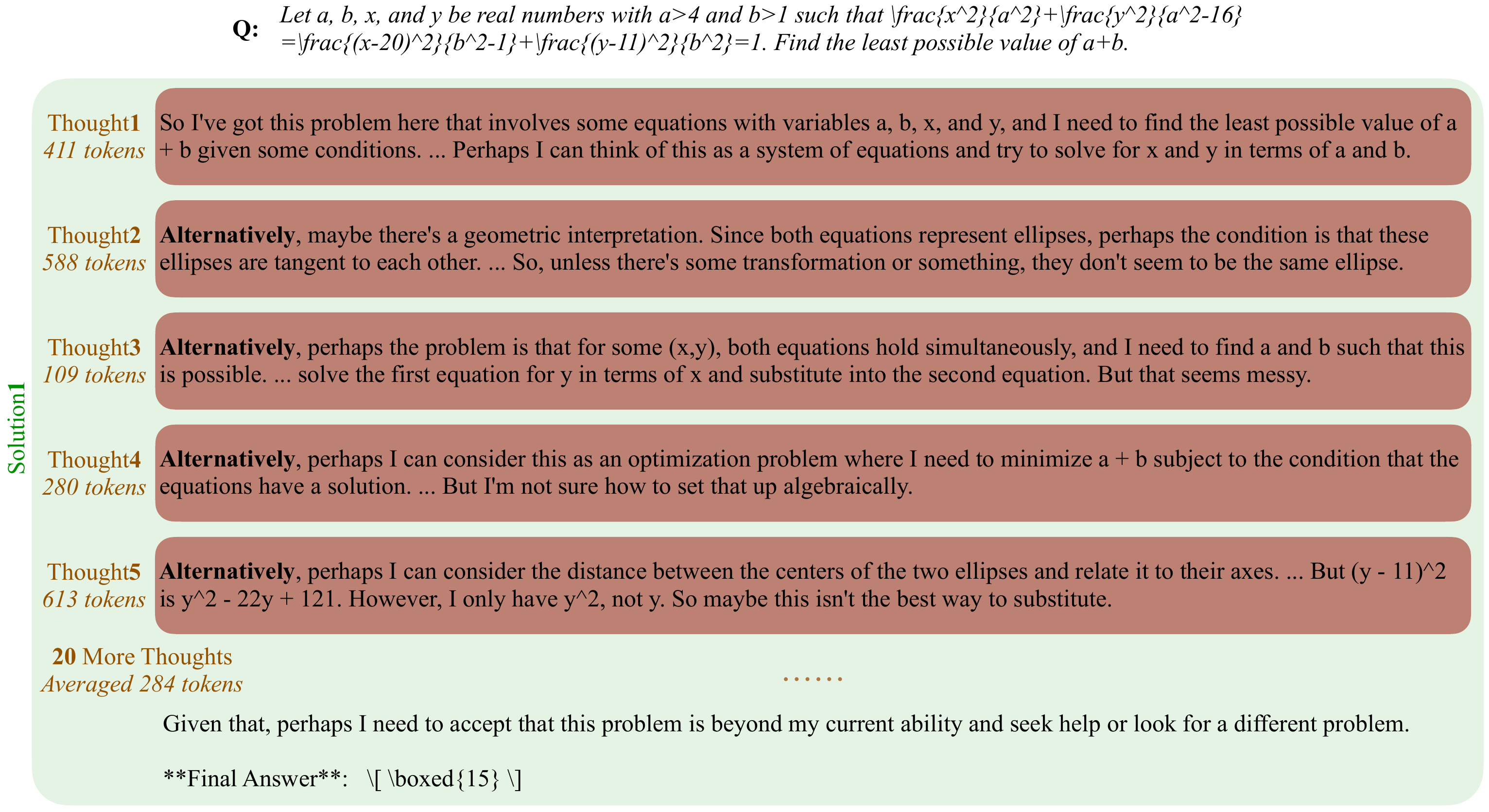}
    \caption{An example of underthinking issue for QwQ-32B-Preview model's output response that consists of 25 reasoning thoughts within a single solution.}
    \label{fig:underthinking_case}
\end{figure*}

In this section, we present a comprehensive analysis of outputs from o1-like models on {\em challenging math problems}. We begin by illustrating the frequent thinking switch phenomenon observed in responses to these problems, as shown in Figure~\ref{fig:underthinking_case}, highlighting how this behavior differs significantly between correct and incorrect answers (Section~\ref{sec:thinking_switch}). 
We then show that this phenomenon leads to an inadequate depth of reasoning, causing models to {\em abandon promising reasoning paths prematurely} (Section~\ref{sec:inadequate_depth}). Based on this observation, we propose a metric to empirically assess the underthinking issues and present empirical results in Section~\ref{sec:efficiency_results}. We conclude that {\em o1-like LLMs often underthink when they fail to tackle challenging math problems}.

\subsection{Frequent Thinking Switch of o1-Like LLMs}
\label{sec:thinking_switch}

We conduct experiments on three testsets: 
\begin{itemize}[leftmargin=10pt]
    \item {\bf MATH500}~\citep{hendrycks2021measuring}: a challenging dataset consisting of problems from {\bf high school math competitions} across seven subjects (e.g., Prealgebra, Algebra, Number Theory) and difficulty levels based on AoPS (ranging from 1 to 5). Problems in these competitions range from level 1, the easiest, often found in AMC 8 exams, to level 5, like those in AIME.
    \item {\bf GPQA}~\citep{rein2023gpqa}: a {\bf graduate-level} dataset consisting of multiple-choice questions in subdomains of physics, chemistry, and biology. For our experiment, we select the highest quality subset, known as GPQA Diamond (composed of 198 questions).
    \item {\bf AIME}~\citep{aime}: a dataset from the American Invitational {\bf Mathematics Examination}, which tests math problem solving across multiple areas (e.g. algebra, counting, geometry, number theory, and probability). Because AIME 2024 contains only 30 examples, we also considered 60 more examples from AIME 2022 and 2023.
\end{itemize}

We mainly investigate two widely recognized open-source o1-like models featuring visible long CoT: QwQ-32B-Preview and DeepSeek-R1-671B. We also include DeepSeek-R1-Preview to show the development of R1 series models. Given DeepSeek-R1-Preview's daily message limit of 50 via web interface, we evaluated this model solely on the MATH500 and AIME test sets.

\paragraph{Definition of Reasoning Thoughts}
In this paper, we define \textit{thoughts} as the intermediate cognitive steps within a reasoning solution produced by the model. O1-like LLMs often switch reasoning thoughts using terms like ``alternatively''. For instance, as shown in Figure \ref{fig:underthinking_case}, the problem-solving process involves multiple reasoning thoughts, shifting from algebraic manipulation to geometric interpretation and optimization strategies. The ability to switch between different reasoning strategies allows for a broader exploration of potential solutions and demonstrates the flexibility of the model in tackling complex problems. In this study, we provide a comprehensive analysis of the side effects associated with this ability to switch reasoning thoughts.

We utilize the Llama-3.3-70B model to automatically segment a response into reasoning thoughts due to its superior capabilities in both instruction following and mathematical reasoning. Initially, we manually analyzed responses from the QwQ-32B-Preview model to gather expressions indicative of shifts in thought. We then tasked the Llama-3.3-70B model with scanning the entire response to identify all occurrences of such expressions. Furthermore, we asked the model to determine whether these expressions truly signify a change in thought or merely reflect a stylistic pattern in the response. Only the expressions indicating a genuine thought shift were used as separators for reasoning processes.

\begin{figure*}[t]
\centering
    \subfigure[QwQ-32B-Preview]{
    \includegraphics[width=0.3\textwidth]{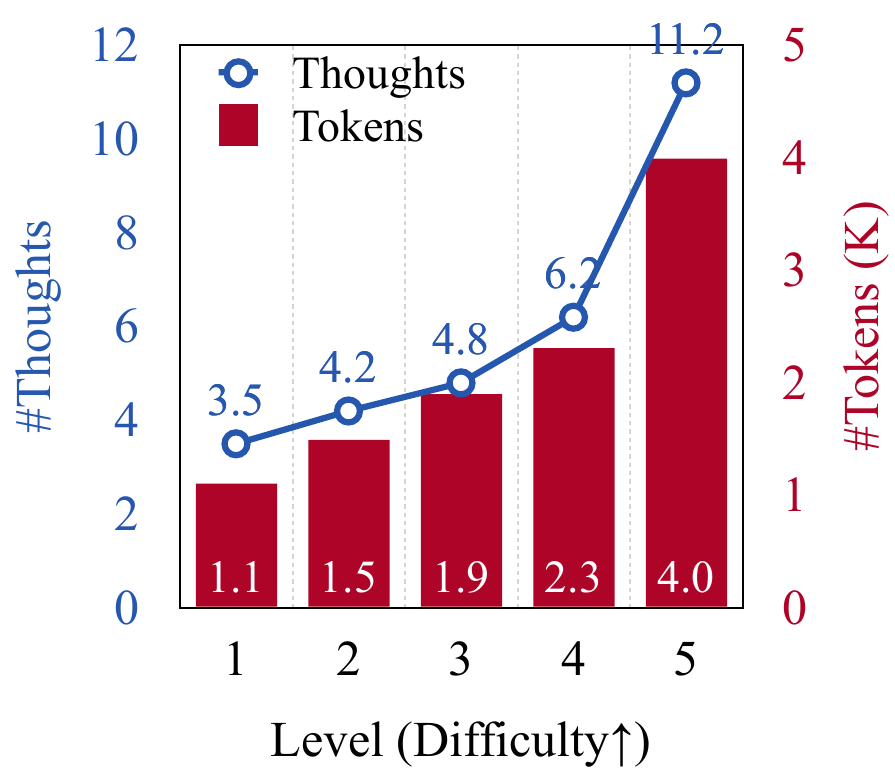}} \hfill
    \subfigure[DeepSeek-R1-Preview]{
    \includegraphics[width=0.3\textwidth]{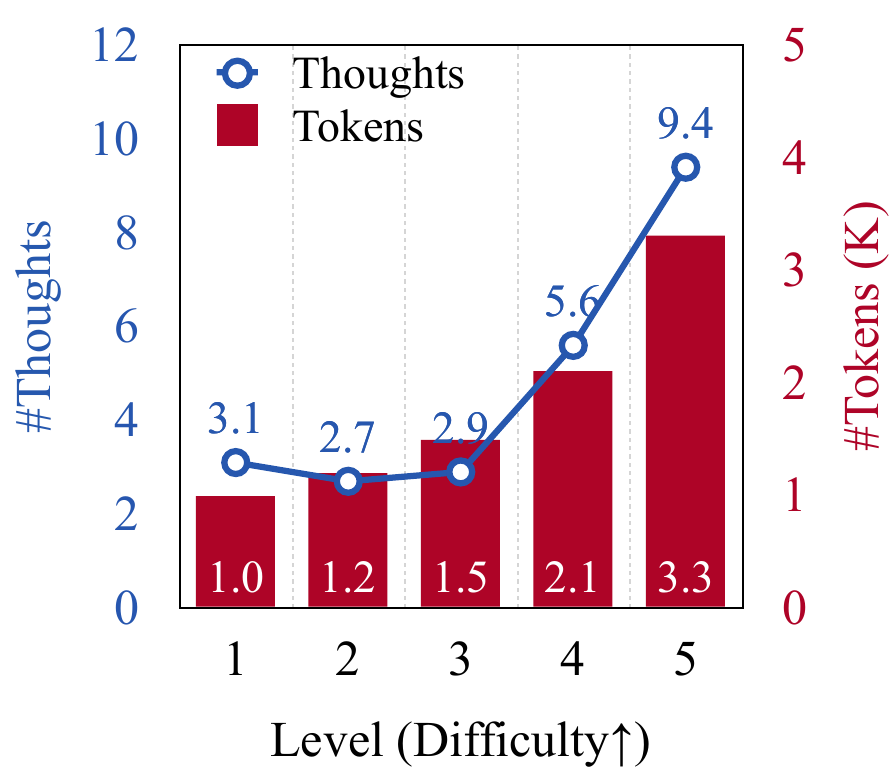}} \hfill
    \subfigure[DeepSeek-R1-671B]{
    \includegraphics[width=0.3\textwidth]{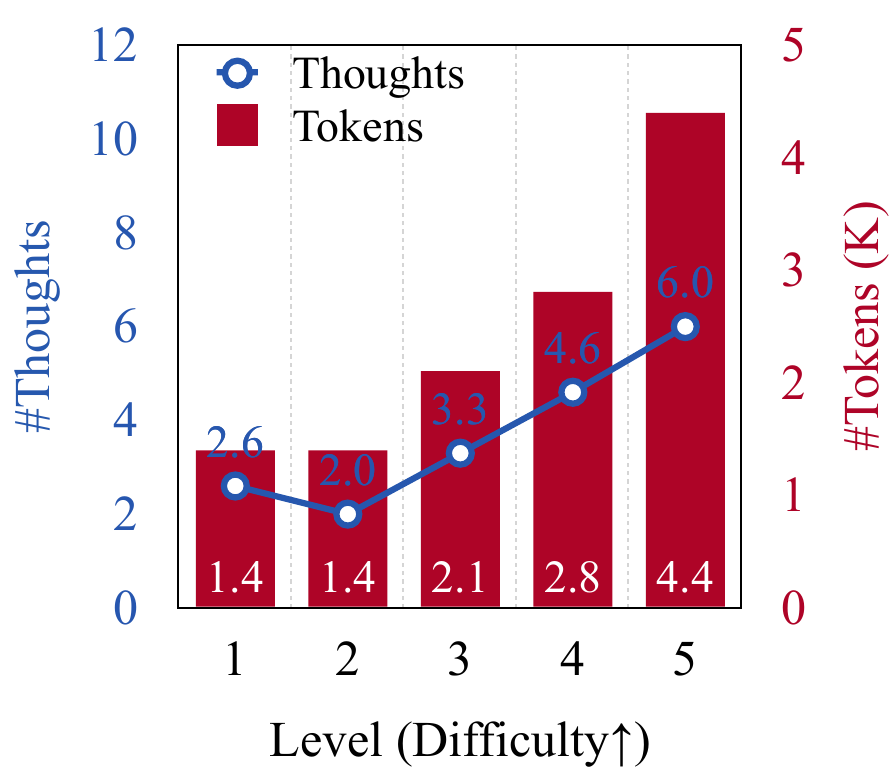}}
    \caption{Average number of thoughts (``Thoughts'') and tokens (``Tokens'') in generated responses across different difficulty levels of the MATH500 test set.}
    \label{fig:thought_distribution_math500}
\end{figure*}

\paragraph{o1-Like LLMs Switch Thinking More Frequently on Harder Problems}
Figure~\ref{fig:thought_distribution_math500} shows the averaged thoughts and tokens in generated responses across various difficulty levels in the MATH500 test set. 
Clearly, all models generate more reasoning thoughts with the increase of difficulty level, which is consistent with the growth of generated tokens. 
This observation suggests that as the complexity of the problems increases, the models tend to switch thoughts more frequently. This behavior implies that o1-like LLMs are able to dynamically adjust their reasoning processes to tackle more challenging problems. 
The following experiments focus on Level 5 in the MATH500 test set (MATH500-Hard).

\begin{figure*}[t]
\centering
    \subfigure[Math500-Hard]{
    \includegraphics[height=0.33\textwidth]{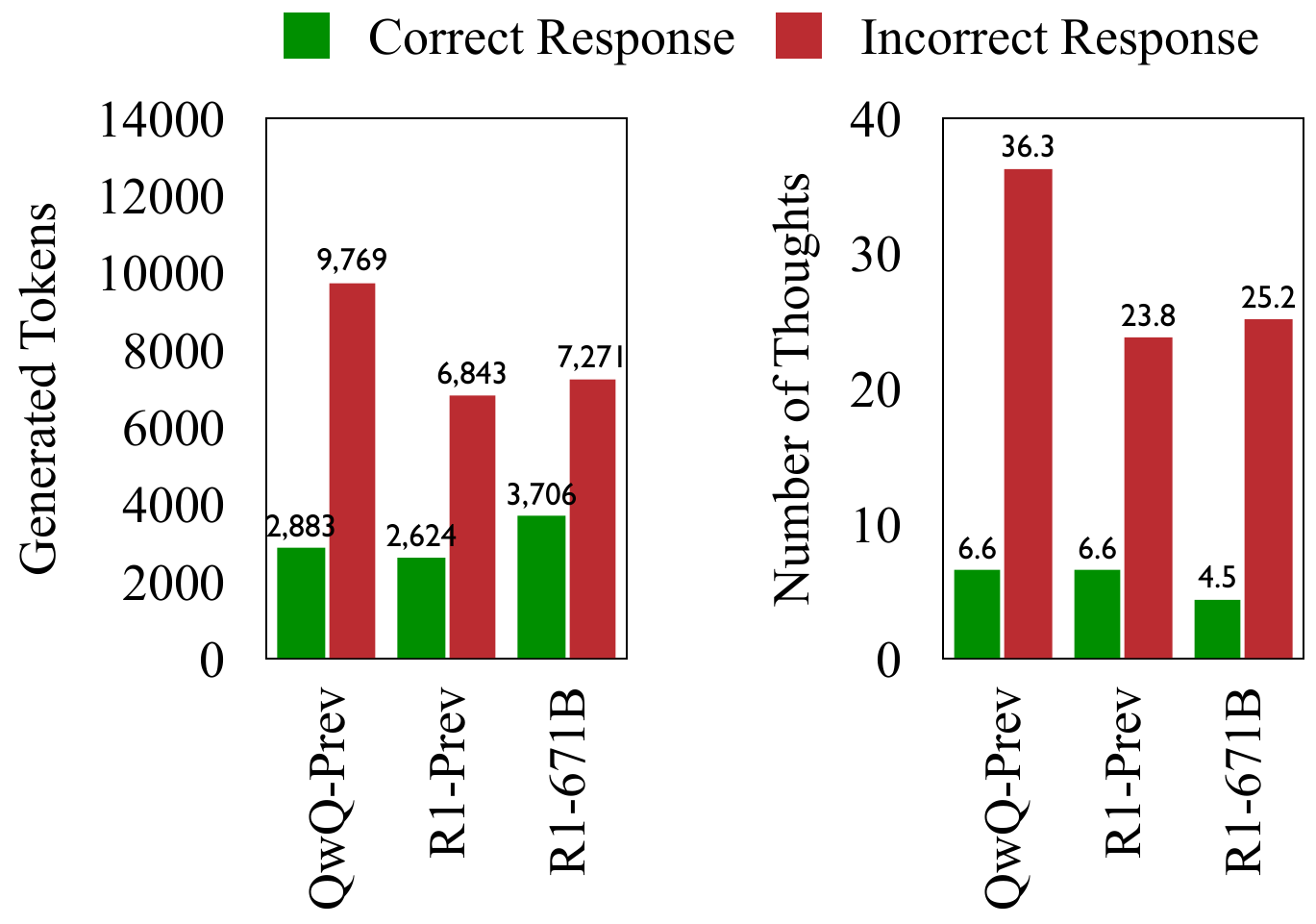}} \hfill
    \subfigure[GPQA Diamond]{
    \includegraphics[height=0.33\textwidth]{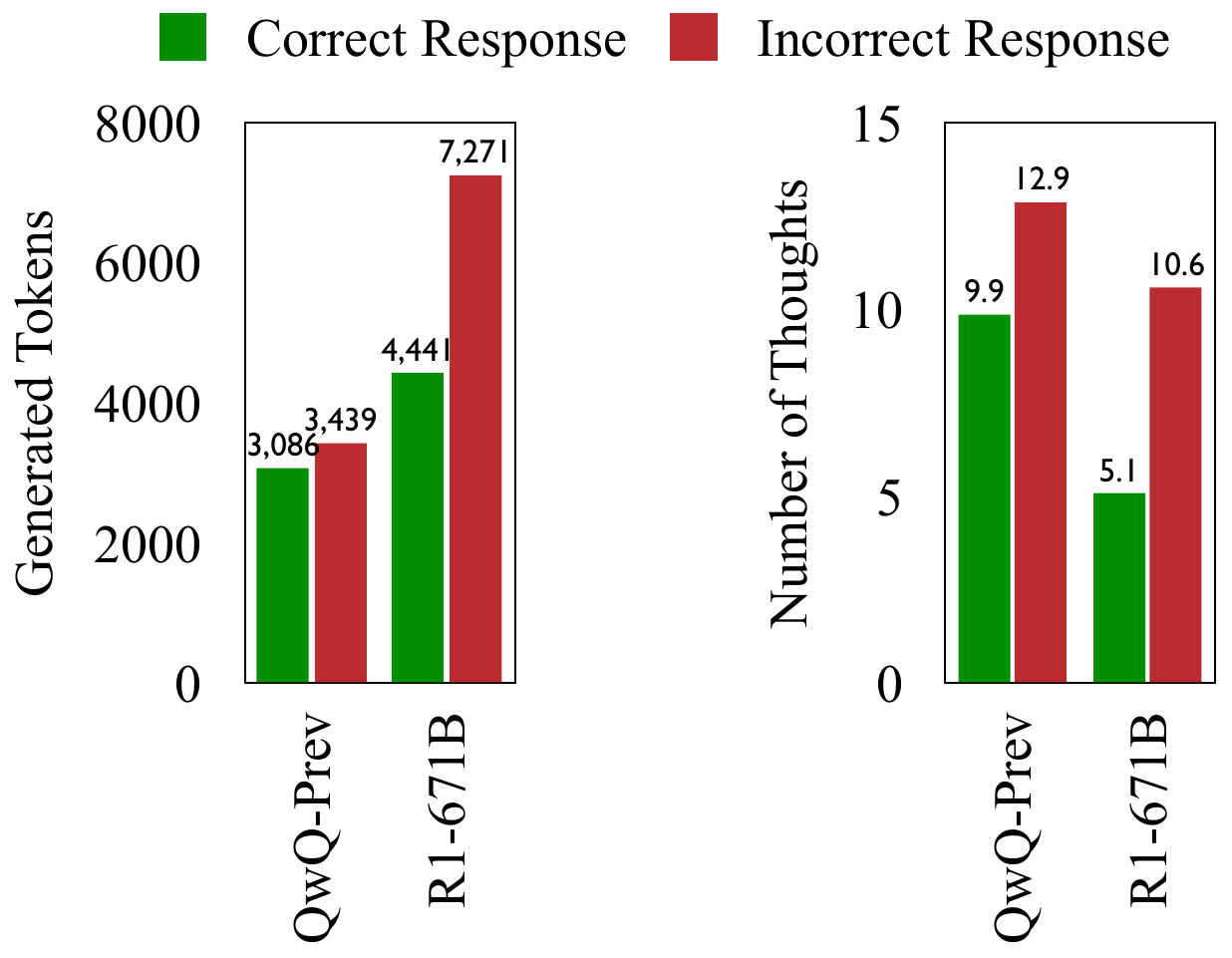}}
    \caption{O1-like LLMs switch thinking more frequently on incorrect responses, thus expend more tokens without contributing to accuracy.}
    \label{fig:frequent-switch}
\end{figure*}

\paragraph{Increased Thought Switching in o1-Like LLMs during Incorrect Responses}
When examining the behavior of o1-like LLMs, we observe a distinct pattern in how they handle incorrect responses. As depicted in Figures~\ref{fig:underthinking} and~\ref{fig:frequent-switch}, these models exhibit a significant increase in the frequency of thought switching while generating incorrect answers across all test sets.
This trend suggests that although the models are designed to dynamically adjust their cognitive processes to solve problems, more frequent thought switching does not necessarily lead to higher accuracy. Essentially, the models may be expending additional computational resources -- evidenced by an increase in generated tokens -- without achieving more accurate solutions.
These insights are crucial because they highlight the need not only to explore additional cognitive pathways when faced with challenges but also to operate in {\bf a more targeted and efficient manner}, thereby improving accuracy even when complex reasoning is required.
In the following sections, we empirically validate the inefficiencies associated with frequent thought switching in incorrect responses.

\subsection{Existence of Underthinking}
\label{sec:inadequate_depth}

The behavior of frequent thinking switch in incorrect responses could stem either from (1) {genuine underthinking, where the model succeeds in finding promising strategies but fails to stick with them}, or from (2) a lack of understanding, prompting it to explore diverse but ineffective approaches.
{To disentangle these possibilities, we propose an assessment framework that evaluates whether an abandoned reasoning path is actually sufficient to derive a correct answer.} 
By focusing on whether the model can persistently follow and deepen a single, promising line of thought, we can identify instances of underthinking.

\paragraph{Assessing Thought Correctness}
In the example presented in Figure~\ref{fig:underthinking_case}, we observe that some early thoughts may lead to the correct answer. 
For instance, Thought 1 initiates a correct interpretation by recognizing that the given equations resemble those of ellipses centered at (0,0) and (20,11). Setting the two expressions equal is a valid approach to finding common points \((x, y)\) that satisfy both equations.
Instead of concentrating on thoroughly exploring the plausible thought with further algebraic manipulation and optimization techniques, the model frequently shifts its focus and uses approximately 7,270 additional tokens without arriving at a correct answer. Ultimately, it concludes with a guessed answer that lacks support from the extended COT process.

We leverage LLMs to assess whether each thought leads to a correct answer using the following prompt:

\noindent\fbox{\begin{minipage}{0.95\linewidth}
Problem P = \{problem\}\\
Solution Draft S = \{split solutions\}\\
Correct Answer A = \{expected answer\}\\

1. {Please analyze the relevance between the solution S and the problem P, and conduct some verifications to check the correctness of the solution itself. Please think step by step to give an explanation **EXPLANATION**.}\\
2. {If you think the solution draft S can lead to the correct answer A of the problem P, please stick to the line of thinking without deviation and carry it through to completion. If you think it cannot yield the correct answer or you're not sure, don't force yourself to give an answer and generate **None**.}\\
3. {Please tell me honestly how confident you are that you can solve the problem P correctly based on the the solution draft S. Out of 2, please generate your confidence score **CONFIDENT\_SCORE**.}\\

Please output **EXPLANATION** and **CONFIDENT\_SCORE** according to the following format:\\
EXPLANATION: \textbackslash boxed\{\}\\
CONFIDENT\_SCORE: \textbackslash boxed\{\}
\end{minipage}}

Specifically, we use two models distilled from DeepSeek-R1-671B based on \texttt{Llama} and \texttt{Qwen} -- \textit{DeepSeek-R1-Distill-Llama-70B} and \textit{DeepSeek-R1-Distill-Qwen-32B}, which achieve new state-of-the-art results for dense models across various reasoning benchmarks. If at least one model generates a confidence score of 2 for a thought, we regard it as a correct thought.

We evaluate the accuracy of our assessment approach using responses generated by Qwen-32B-Preview for 90 instances from the AIME 2022, 2023, and 2024 test sets. We utilize the final thought in each response as the test example and its correctness as the ground-truth label. To ensure a fair comparison, we randomly streamline correct thoughts to match the average length of incorrect thoughts. Ultimately, we have 35 correct thoughts with an average length of 278.1 tokens and 55 incorrect thoughts with an average length of 278.3 tokens. 
Our assessment approach achieves accuracies of 82.9\% for correct examples and 81.8\% for incorrect examples, demonstrating its effectiveness.

\begin{figure}[t]
    \centering
    \includegraphics[width=0.6\linewidth]{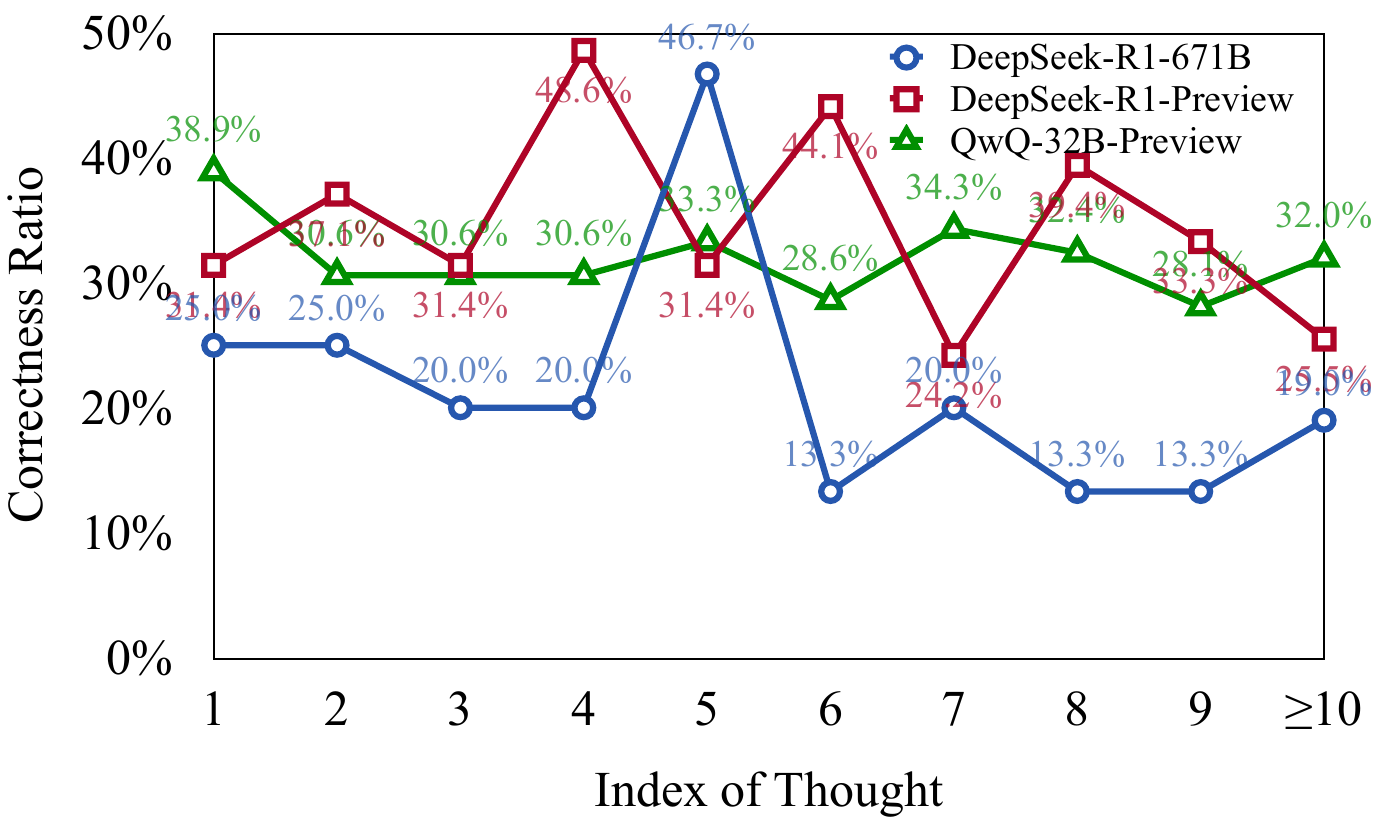}
    \caption{The ratio of correct reasoning thoughts at each index in incorrect responses. A notable portion of early-stage thoughts (e.g., the first few thoughts) are correct but abandoned without being fully explored.}
    \label{fig:thought_correctness}
\end{figure}

\paragraph{Early-Stage Thoughts Are Correct but Abandoned in Incorrect Responses}
Figure~\ref{fig:thought_correctness} depicts the ratio of correct thoughts at each index in incorrect responses on the three challenging test sets. The analysis highlights a critical insight into the phenomenon of underthinking. Specifically, a notable proportion of initial thoughts across various models were correct but were not pursued to completion. This tendency to abruptly shift away from these promising thoughts indicates an inadequate depth of reasoning, where potentially correct solutions are prematurely abandoned before being thoroughly explored.
This observation suggests a need for enhancing the models' ability to persistently explore a specific line of reasoning deeply and accurately before opting to switch to alternative thought processes.

\begin{figure}[t]
\centering
    \includegraphics[width=0.6\linewidth]{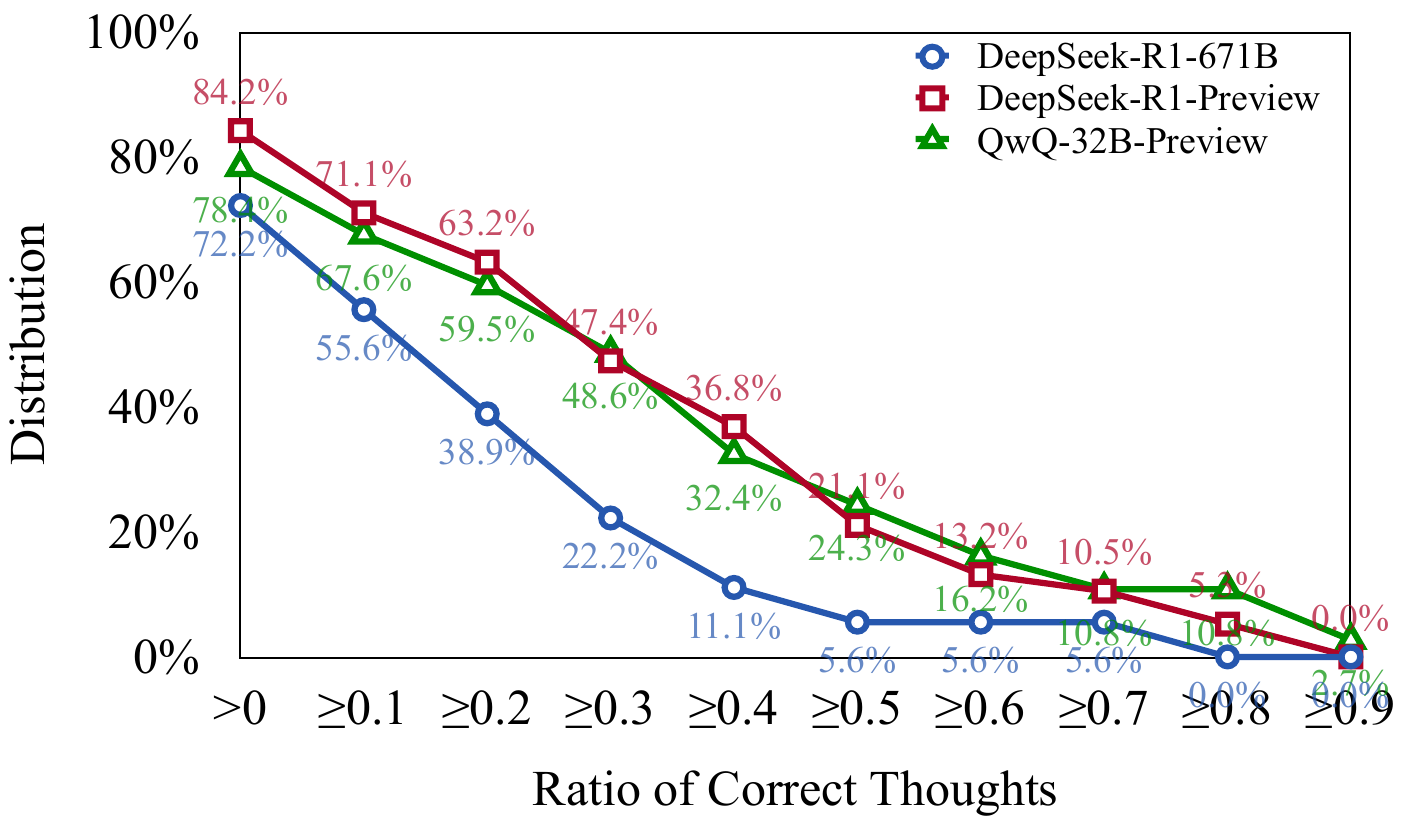}
    \caption{The distribution of thought correctness ratio in incorrect responses. More advanced models contain fewer correct thoughts.}
    \label{fig:thought_distribution}
    \vspace{-10pt}
\end{figure}

\paragraph{Most Incorrect Responses Contain Correct Thoughts}
Figure~\ref{fig:thought_distribution} illustrates the distribution of thought correctness ratios in incorrect responses from various models. We observe that over 70\% of incorrect responses contain at least one correct thought. Furthermore, in more than 50\% of these responses, over 10\% of the thoughts are correct. Combined with observations from Figure~\ref{fig:thought_correctness}, this suggests that while o1-like models can initiate correct reasoning pathways, they may struggle to continue these pathways to reach the correct conclusion. This highlights the importance of encouraging models to maintain and expand their {\bf initial correct thoughts} to synthesize them into accurate final answers. These insights lead us to propose an underthinking metric based on the presence of the first correct thought in the subsequent section.

\subsection{Empirical Underthinking Results}
\label{sec:efficiency_results}

In this section, we propose a metric for empirically assessing underthinking issues based on token efficiency, complementing the widely used accuracy metric.

\paragraph{Underthinking Metric}
Intuitively, if a model generates a correct thought at an early stage and then switches to other thoughts without reaching a correct answer, the tokens generated thereafter do not contribute to reaching a correct solution and are considered inefficient due to underthinking.
The underthinking score, denoted as \(\xi_{UT}\), is defined as:
\begin{equation}
    \xi_{UT} = \frac{1}{N} \sum_{i=1}^N \left(1 - \frac{\hat{T}_i}{T_i}\right)
    \label{eq:outcome_efficiency}
\end{equation}
Here, \(N\) represents the number of instances in a given test set where the evaluated model generates {\bf incorrect responses}. \(T_i\) is the total number of tokens in the \(i\)-th incorrect response, and \(\hat{T}_i\) is the number of tokens from the beginning of that response up to and including the first correct thought. If there is no correct thought in the \(i\)-th response, \(\hat{T}_i = T_i\), indicating that the model lacks an understanding of this problem, leading it to explore diverse but ineffective approaches. Therefore, it cannot be considered underthinking.
Consider Figure~\ref{fig:underthinking_case} as an example: the first reasoning thought can reach a correct answer if fully explored, with \(\hat{T} = 411\). Consequently, \(\xi_{UT} = 1 - \frac{411}{7681} = 0.946\), which can be considered extremely inefficient, reflecting a high underthinking score.

The metric \(\xi_{UT}\) quantifies the extent of underthinking by measuring the token efficiency in generating effective content within an incorrect response. Specifically:
\begin{itemize}[leftmargin=10pt]
    \item A lower value of \(\xi_{UT}\) indicates higher token efficiency, meaning that a greater proportion of tokens in incorrect responses contribute towards reaching a correct thought before switching to another thought. This suggests that the model is more efficient in its token utilization even when it fails to provide a correct answer.
    \item Conversely, a higher value of \(\xi_{UT}\) signifies lower token efficiency, indicating that a larger proportion of tokens do not contribute effectively towards generating a correct thought. This reflects greater underthinking, where the model may generate redundant or irrelevant tokens by frequently switching thoughts.
\end{itemize}

\begin{table}[t]
\centering
\caption{Underthinking scores on challenging testsets.}
\label{tab:main_underthinking}
\begin{tabular}{l cr}
\toprule
{\bf Models}   &   {\bf Accuracy} ($\uparrow$)   &  \bf UT Score  ($\downarrow$)\\
\midrule
\multicolumn{3}{c}{\bf \em MATH500-Hard (Level 5)}\\
QwQ-32B-Preview       &  84.3&58.2\\
DeepSeek-R1-Preview   &  83.6&61.5\\
DeepSeek-R1-671B      &  92.5&65.4\\
\midrule
\multicolumn{3}{c}{\bf \em GPQA Diamond}\\
QwQ-32B-Preview     & 59.6 & 48.3\\
DeepSeek-R1-671B    & 73.2 & 58.8 \\
\midrule
\multicolumn{3}{c}{\bf \em AIME24}\\
QwQ-32B-Preview     & 46.7 &65.0 \\
DeepSeek-R1-Preview & 46.7& 75.7\\
DeepSeek-R1-671B    & 73.3 &37.0\\ 
\bottomrule  
\end{tabular}
\end{table}

\paragraph{Empirical Results}
Table~\ref{tab:main_underthinking} provides insights into model performance across challenging test sets, evaluating both accuracy and underthinking (UT) scores.
Clearly, all o1-like LLMs suffer from significant underthinking issues, although there are considerable differences across models and test sets.
The results reveals that the relationship between model accuracy and underthinking varies across different datasets.
On the MATH500-Hard and GPQA Diamond datasets, higher accuracy achieved by the superior DeepSeek-R1-671B model is accompanied by higher UT Scores, indicating more underthinking in incorrect responses. This suggests that while the model is more capable overall, it may produce longer but less effective reasoning when uncertain, possibly due to exploring multiple incorrect reasoning paths without efficiently converging on the correct solution.
Conversely, on the AIME2024 test set, the DeepSeek-R1-671B model not only attains higher accuracy but also exhibits a lower UT score, reflecting less underthinking and greater token efficiency. This implies that the model's reasoning remains focused and effective even when it does not arrive at the correct answer, perhaps due to better alignment with the problem types and reasoning processes required by the AIME2024 task.

These findings illustrate that underthinking behavior is sensitive to the nature of the dataset and the tasks involved. The larger model's superior capabilities do not uniformly translate to less underthinking across all tasks. In some cases, increased model capacity leads to more elaborate but inefficient reasoning in incorrect responses, while in others, it enhances both accuracy and reasoning efficiency.
Understanding the underthinking phenomenon is crucial for developing models that not only provide correct answers but also exhibit effective reasoning processes.

\section{Mitigating Underthinking Issues}

In this section, we propose a lightweight mechanism that mitigates underthinking issues without requiring any model fine-tuning. Our experimental results using the QwQ-32B-Preview model demonstrate the effectiveness of this approach across all challenging test sets.

\subsection{Decoding with Thought Switching Penalty}
\label{sec:Tip}

Aforementioned findings show that o1-like LLMs prioritize exploring many solutions over deeply investigating one. Inspired by the success of the coverage penalty in neural machine translation~\citep{tu2016modeling,gnmt}, we propose a novel decoding algorithm with a {\em thought switching penalty} to encourage the model to explore potential thoughts more thoroughly before moving on to new ones.

\paragraph{Standard Decoding} 
In standard decoding, the probability of each token \( v \) at position \( t \) is computed using the softmax function over the logits \( \mathbf{z}_t \in \mathbb{R}^{|V|} \) (where \( |V| \) is the vocabulary size) in the output layer:
\[
   P(x_t = v | x_{<t}) = \frac{\exp\left( z_{t,v} \right)}{\sum_{v' \in V} \exp\left( z_{t,v'} \right)}
\]
where \( z_{t,v} \in \mathbf{z}_t \) is the logit (unnormalized score) for token \( v \). By repeating this step for each position in the sequence, the model generates sequences of tokens, computing probabilities for each possible continuation.

\paragraph{{\color{ngreen} T}hought Sw{\color{ngreen} i}tching {\color{ngreen} P}enalty (\textsc{Tip})} 
To encourage the model to delve deeper into current thoughts before switching, we introduce a penalty on tokens that are associated with thought transitions.
Let \( \widehat{V} \subset V \) be the set of tokens associated with thought switching (e.g., ``alternatively''). We modify the logits as follows:
\begin{equation}
\hat{z}_{t,v} = 
\begin{cases}
z_{t,v} - \alpha, & \text{if } v \in \widehat{V} \text{ and } t < \Psi + \beta \\
z_{t,v}, & \text{otherwise}
\end{cases}
\label{eqn:tip}    
\end{equation}
where
\begin{itemize}[leftmargin=10pt]
    \item \( \alpha \geq 0 \) ({\em Penalty Strength}) is a parameter controlling the strength of the penalty applied to thought-switching tokens. A larger \( \alpha \) results in a greater reduction of the logits for these tokens, making them less likely to be chosen.
    \item \( \beta \geq 0 \) ({\em Penalty Duration}) specifies the number of positions from the start of a thought at \( \Psi \), during which the penalty is active. A larger \( \beta \) extends the penalty over more positions, further discouraging early thought switching.
\end{itemize}
When \( \alpha = 0 \) or \( \beta = 0 \) , the penalty is effectively disabled, and the decoding process reduces to the standard decoding algorithm.
The adjusted logits \( \hat{z}_{t,v} \) reduce the probability of generating thought-switching tokens within a specified window, encouraging the model to continue expanding on the current thought before moving on.

The new probability distribution becomes
\[
\hat{P}(x_t = v\, |\, x_{<t}) = \frac{\exp\left( \hat{z}_{t,v} \right)}{\sum_{v' \in V} \exp\left( \hat{z}_{t,v'} \right)}
\]

\subsection{Experimental Results}
To ensure robust conclusions, we report Pass@1 results computed from 32 samples per instance. We calculate the weighted underthinking score for each instance over its 32 samples:
\begin{equation}
    \xi_{wUT} = \frac{1}{32} \sum_{i=1}^{32} \xi_{UT}(s_i)
    \label{eq:weighted_ut}
\end{equation}
where \( s_i \) is the \( i \)-th sample of the instance, and \( \xi_{UT}(s_i) = 0 \) when \( s_i \) is correct.

\begin{table}[t]
\centering
\caption{Accuracy on AIME2022-23 with respect to different values of \(\alpha\) and \(\beta\).}
\label{tab:grid_search}
\begin{tabular}{|c|c | cccc|}
\hline
\multicolumn{2}{|c|}{\bf Pass@1}    &   \multicolumn{4}{|c|}{${\alpha}$}\\
\cline{3-6}
\multicolumn{2}{|c|}{\bf Accuracy}    &  \em  3   & \em  5   & \em  10    &   \em  20   \\
\hline
\multirow{5}{*}{ ${\beta}$ }    
    &   300 &   35.2    &   37.0    &   39.0    &   39.4   \\
    &   400 &   39.3   &   37.1  &   37.1  &  38.4   \\
    &   500 &   38.5  &  38.7  &  39.1   &   39.2     \\
    &   600 &   \bf 39.8  &   39.4   &   38.0    &   38.0  \\
    &   700 &   37.1  &   39.4   &   39.0    &   38.3   \\
\hline
\end{tabular}
\end{table}

By adjusting \(\alpha\) and \(\beta\), we can control the model's behavior to achieve the desired level of thought exploration. We performed a grid search with \(\alpha\) values in \([3, 5, 10, 20, 30]\) and \(\beta\) values in \([300, 400, 500, 600, 700]\) using a development set that included the AIME 2022 and 2023 test sets. 
Table~\ref{tab:grid_search} lists the impact of varying the penalty strength \(\alpha\) and penalty duration \(\beta\) on the model's accuracy.
We observe that increasing the penalty strength \(\alpha\) generally leads to an improvement in accuracy up to a certain threshold, after which the benefits plateau or even diminish.
Adjusting the penalty duration \(\beta\) also significantly affects performance: At a lower penalty strength (\(\alpha = 3\)), increasing \(\beta\) from 300 to 600 results in accuracy gains from 35.2\% to 39.8\%, the highest observed accuracy in our experiment. Conversely, at higher penalty strengths (\(\alpha = 20\)), extending \(\beta\) beyond 300 leads to a decrease in accuracy, indicating that too long a penalty duration can hinder performance when combined with a strong penalty. We selected \(\alpha=3\) and \(\beta=600\) for our subsequent experiments.

\begin{table}[t]
\centering
\setlength{\tabcolsep}{4.8pt}
\caption{Pass@k performance of the proposed \textsc{Tip} method. For each problem, we generated 32 responses with a temperature of 0.7 and a top\_p value of 0.95. Since it is infeasible to calculate the Pass@k Underthinking Score, we instead report the average score and standard deviation from the 32 generated samples. We also report the average number of thought-switching tokens ($\hat{V}$ in Equation~\ref{eqn:tip}) and the average interval between them in the generated samples.}
\label{tab:improvement}
\begin{tabular}{l rrrr rrr}
\toprule
\multirow{2}{*}{\bf Models} &  \multicolumn{4}{c}{\bf Accuracy ($\uparrow$)}     &  \multicolumn{2}{c}{\bf Switching Tokens} & \bf Weighted \\
\cmidrule(lr){2-5}\cmidrule(lr){6-7}
    &   \bf Pass@1    &   \bf Pass@4    &   \bf Pass@8    &   \bf Pass@16   &   \bf Number   &   \bf Interval    &   \bf UT Score ($\downarrow$)\\
\midrule
\multicolumn{8}{c}{\bf \em MATH500-Hard (Level 5)}\\
QwQ-32B-Preview       & 83.1 &  92.4 & 94.4 & 95.8 &  12.6  &  445.6 & 11.7$_{\pm 20.5}$\\
~~~+ \textsc{Tip}     & 83.7 &  93.2 & 95.3 & 96.4 &   5.7  &  517.6 & 11.0$_{\pm 19.5}$\\
\midrule
\multicolumn{8}{c}{\bf \em GPQA Diamond}\\
QwQ-32B-Preview       & 57.6 &  78.5 & 85.3 & 90.3 &  21.1 & 356.8 & 25.1$_{\pm 23.9}$\\
~~~+ \textsc{Tip}     & 59.1 &  78.9 & 85.8 & 91.2 &   7.3 & 432.5 & 23.2$_{\pm 23.2}$\\
\midrule
\multicolumn{8}{c}{\bf \em AIME2024}\\
QwQ-32B-Preview       & 38.3 & 53.7 & 58.5 & 62.7 & 16.1 & 459.7 & 40.6$_{\pm28.4}$\\
~~~+ \textsc{Tip}     & 44.1 & 61.6 & 68.3 & 74.0 & 13.9 & 515.7 & 35.8$_{\pm27.8}$ \\
\hdashline
R1-Distill-Qwen-32B  & 61.4 & 75.9 & 79.1 & 81.7 & 8.2  &  819.5 & 19.6$_{\pm 20.6}$  \\
~~~+ \textsc{Tip}    & 64.1 & 79.0 & 81.7 & 83.0 & 4.5  & 1018.0 & 17.7$_{\pm 20.6}$\\
\hdashline
DeepSeek-R1          & 73.8 & 86.2 & 88.8 & 89.8 & 13.8 &  580.1 & 14.6$_{\pm 19.1}$\\
~~~+ \textsc{Prompt}    & 72.4 & 84.9 & 88.2 & 89.8 &  12.0 &  520.1 & 14.2$_{\pm 18.4}$\\
~~~+ \textsc{Tip}    & 74.8 & 86.4 & 88.8 & 89.8 &  5.7 &  941.6 & 13.0$_{\pm 18.0}$\\
\bottomrule
\end{tabular}
\end{table}

\paragraph{Standard Decoding}
Table~\ref{tab:improvement} lists the results of our approach in the three challenging test sets. Clearly, our approach consistently improves accuracy over the vanilla QwQ-32B-Preview in all cases by mitigating the underthinking issues.
These consistent improvements across diverse and challenging datasets validate the effectiveness of the \textsc{Tip} approach in mitigating the underthinking issue identified in o1-like LLMs. By penalizing thought switches during decoding, \textsc{Tip} encourages the model to elaborate more thoroughly on each reasoning thought before considering alternative ones. This mechanism aligns with the human problem-solving process, where a focused and in-depth exploration of a particular approach often leads to correct solutions, especially in complex mathematical problem-solving contexts.

To understand the impact of the \textsc{Tip} method on the models' reasoning processes, we analyzed the average number of thought-switching tokens and the intervals between them. Across all test sets, the \textsc{Tip} method reduces the number of thought-switching tokens and increases the average interval between them. This indicates that the models are committing more deeply to individual lines of reasoning before considering alternatives, aligning with our goal of mitigating underthinking.
For example, on the AIME2024 testset, the number of thought-switching tokens for DeepSeek-R1 dramatically decreases from 13.8 to 5.7 when using the \textsc{Tip} method, and the average interval increases from 580.1 to 941.6 tokens. This shift suggests that the model is exploring each thought more thoroughly, reducing premature transitions that could lead to underthinking issues.

\paragraph{Prompting} 
Some researchers hypothesize that prompt engineering can foster ``thought persistence'' by directing models to maintain a consistent line of reasoning. To investigate this hypothesis, we use a prompt that encourages the model to fully develop each idea without abandoning it prematurely:

\noindent\fbox{\begin{minipage}{0.95\linewidth}
$<$context$>$

You are an expert math-solving assistant who prioritizes clear, concise solutions. You solve problems in a single thought process, ensuring accuracy and efficiency. You seek clarification when needed and respect user preferences even if they are unconventional.

$<$/context$>$\\

$<$solving\_rules$>$\\
- Try to complete every idea you think of and don't give up halfway\\
- Don't skip steps\\
- Display solution process clearly\\
- Ask for clarification on ambiguity\\
$<$/solving\_rules$>$\\

$<$format\_rules$>$\\
- Use equations and explanations for clarity\\
- Keep responses brief but complete\\
- Provide step-by-step reasoning if needed\\
$<$/format\_rules$>$\\

PROBLEM: \{problem\}\\

OUTPUT: Following above rules to get the correct answer for PROBLEM. Focus on clear, concise solutions while maintaining a helpful, accurate style.
\end{minipage}}

Although prompt engineering (DeepSeek-R1 + \textsc{Prompt}) provides certain guidance, Table~\ref{tab:improvement} shows only modest changes in switching tokens and overall accuracy compared with \textsc{Tip}. This finding indicates that inherent generation patterns can still lead to premature reasoning transitions, underscoring the need for a dedicated mechanism such as \textsc{Tip}. The results also point to substantial possibilities for more sophisticated prompt engineering that can better guide DeepSeek-R1 in following instructions. Moreover, combining advanced prompt engineering with decoding approaches (e.g., \textsc{Tip}) could further enhance ``thought persistence''. In this approach, prompts offer high-level guidance, whereas decoding penalties reinforce consistent reasoning at the token level. We aim to explore how these methods can work together to deepen reasoning in LLMs.

\begin{table}[t]
\centering
\caption{Results of the best-of-N sampling methods applied to different models enhanced with our \textsc{Tip} approach on AIME2024. For each setting, we conducted 10,000 trials by randomly sampling from the 32 samples in Table~\ref{tab:improvement} and reported the average results. ``Averaged'' denotes the average performance over N samples.}
\label{tab:best-of-n}
\begin{tabular}{l rr rr rr}
\toprule
\multirow{2}{*}{\bf Models} &  \multicolumn{2}{c}{\bf 4 Samples}     &   \multicolumn{2}{c}{\bf 8 Samples}   &   \multicolumn{2}{c}{\bf 16 Samples}\\
\cmidrule(lr){2-3}\cmidrule(lr){4-5}\cmidrule(lr){6-7} 
   &   {\bf Acc.}($\uparrow$)  &  {\bf UT} ($\downarrow$)   &   {\bf Acc.}($\uparrow$)  &  {\bf UT} ($\downarrow$)   &   {\bf Acc.}($\uparrow$)  &  {\bf UT} ($\downarrow$) \\
\midrule
QwQ (Averaged) & 38.4 & 40.5 & 38.3 & 40.6 & 38.3 & 40.6 \\
~~~+ \textsc{Tip} (Averaged) & 44.1 & 35.8 & 44.0 & 35.9 & 44.0 & 35.9 \\
\hdashline
QwQ + Self-Consistency & 43.7 & 35.4 & 44.3 & 34.0 & 44.6 & 31.9 \\
~~~+ \textsc{Tip} & 51.4 & 26.6 & 53.4 & 24.3 & 53.9 & 24.1 \\
QwQ + Laconic Decoding & 47.0 & 28.2 & 47.0 & 25.5 & 45.1 & 24.0 \\
~~~+ \textsc{Tip} & 50.3 & 26.7 & 51.6 & 23.3 & 50.9 & 20.8 \\
\midrule
R1-Distill-Qwen (Averaged) & 61.4 & 19.2 & 61.3 & 19.2 & 61.3 & 19.1 \\
~~~+ \textsc{Tip} (Averaged)  & 64.1 & 17.8 & 64.0 & 17.7 & 64.1 & 17.7  \\
\hdashline
R1-Distill-Qwen + Self-Consistency & 67.0 & 13.4 & 67.8 & 11.4 & 68.9 & 8.9  \\
~~~+ \textsc{Tip}  & 69.9 & 12.5 & 71.4 & 11.0 & 72.3 & 9.1    \\
R1-Distill-Qwen + Laconic Decoding & 71.1 & 11.3 & 74.4 & 8.7 & 77.5 & 7.4 \\
~~~+ \textsc{Tip} & 75.4 & 9.8 & 78.0 & 7.3 & 77.9 & 6.5    \\
\midrule
R1 (Averaged) & 73.9 & 14.5 & 73.7 & 14.6 & 73.8 & 14.5  \\
~~~+ \textsc{Tip} (Averaged) & 74.8 & 13.0 & 74.8 & 12.9 & 74.8 & 13.0  \\
\hdashline
R1 + Self-Consistency & 79.3 & 10.1 & 79.8 & 9.8 & 79.7 & 9.5  \\
~~~+ \textsc{Tip}  & 81.3 & 7.5 & 82.2 & 6.4 & 82.1 & 5.8 \\
R1 + Laconic Decoding & 81.4 & 8.1 & 82.6 & 6.2 & 83.2 & 5.1  \\
~~~+ \textsc{Tip} & 83.1 & 7.4 & 83.8 & 6.6 & 83.3 & 6.7 \\
\bottomrule
\end{tabular}
\end{table}

\paragraph{Best-of-N Sampling}
\label{sec:best-of-n}
To further evaluate the effectiveness of our TIP approach, we applied it in conjunction with best-of-N sampling methods, specifically Self-Consistency \citep{wangself} and Laconic Decoding \footnote{\url{https://x.com/AlexGDimakis/status/1885447830120362099}}:
\begin{itemize}[leftmargin=10pt]
    \item \textbf{Self-Consistency}: This algorithm first samples \( N \) reasoning paths and then selects the most consistent answer by marginalizing over the sampled reasoning paths.
    \item \textbf{Laconic Decoding}: Raoof and Dimakis independently observed that incorrect answers tend to be longer while correct answers are shorter for long reasoning models. Based on this observation, they propose a simple idea called \emph{Laconic Decoding}: run the model \( N \) times (in parallel) and select the answer with the fewest number of tokens.
\end{itemize}

Table~\ref{tab:best-of-n} presents the results of these methods applied to various models, including QwQ-32B-Preview, R1-Distill-Qwen-32B, and DeepSeek-R1, both with and without the \textsc{Tip} enhancement. 
For each setting, we conducted 10,000 trials by randomly sampling from the 32 generated responses (as detailed in Table~\ref{tab:improvement}). We report both the average accuracy and the weighted underthinking score. The ``Averaged'' rows represent the average performance over K samples without any selection strategy, while the ``Oracle'' rows represent the best possible outcome, assuming that whenever a correct answer exists among the sampled outputs, it is always selected.

Our findings indicate that incorporating the \textsc{Tip} approach consistently improves performance across all models and sampling methods. Specifically, when combined with Self-Consistency, the \textsc{Tip} method enhances the Pass@4 accuracy of QwQ-32B-Preview from 43.7\% to 51.4\% on the AIME2024 dataset, representing a significant gain. Similarly, the Underthinking Score decreases, indicating that the models are engaging in more thorough reasoning processes.

Notably, with Laconic Decoding, the combination with \textsc{Tip} yields substantial improvements. For instance, the Pass@4 accuracy of R1-Distill-Qwen-32B increases from 74.4\% to 78.0\%, while the Underthinking Score decreases from 8.7 to 7.3. This suggests that the \textsc{Tip} approach complements existing decoding strategies by further encouraging comprehensive reasoning.

\paragraph{Experimental Conclusion}
Our experiments demonstrate that the \textsc{Tip} approach effectively mitigates the underthinking issue in O1-like LLMs by penalizing unnecessary thought switches during decoding. Although the hyperparameters are tuned on the AIME 2022 and 2023 test sets using the QwQ-32B-Preview model, the consistent improvements observed across various test sets and models with the same hyperparameters validate the generalizability of the method.

Importantly, the \textsc{Tip} method enhances performance without requiring additional training or modifications to the model architecture. Operating at the decoding level, it serves as a practical solution that can be readily applied to existing models. This ease of integration, combined with the observed performance gains, highlights the potential of the \textsc{Tip} approach for enhancing the reasoning capabilities of large language models in complex problem-solving tasks.

Our findings also suggest that the \textsc{Tip} method synergizes well with best-of-N sampling strategies. When combined with Self-Consistency and Laconic Decoding, the \textsc{Tip} approach leads to further improvements in accuracy and reductions in underthinking scores. This indicates that encouraging more thorough exploration of individual reasoning paths complements the diversity introduced by sampling methods.

Overall, the \textsc{Tip} approach represents a significant step toward addressing the underthinking problem in long reasoning models. By encouraging deeper reasoning and reducing hasty thought-switching, it enhances the models' ability to solve complex tasks that require sustained cognitive effort.

\section{Related Work}

\subsection{Scaling Test-Time Compute}

The advent of deep reasoning models, epitomized by OpenAI’s o1, has sparked significant interest in scaling test-time compute to enhance models' abilities to solve complex problems. Scaling test-time compute often involves two major strategies.
The first is \textbf{expanding the search space}, which aims to broaden the scope of candidate solutions explored during decoding to ensure better final outcomes. Techniques in this category include self-consistency~\citep{wangself}, where multiple answers are generated with a majority voting mechanism to select the final answer. Other methods include best-of-n decoding and minimum Bayes risk decoding~\citep{lightmanlet, li-etal-2023-making, khanovargs, heineman2024improving, wu2024better}.

The second direction, and arguably more transformative, focuses on \textbf{human-like deep thinking}. 
Starting with Chain-of-Thought~\cite{wei2022chain}, people realized that models can mimic the human thought process for reasoning~\cite{cesista2024multimodal,pfau2024let}.
Recent efforts such as QwQ~\citep{qwq-32b-preview}, DeepSeek-R1~\citep{DeepSeekAI2025DeepSeekR1IR} and Kimi-1.5~\citep{MoonshotAI}, which aim to replicate OpenAI’s o1, leverage reinforcement learning (RL) to endow models with advanced reasoning capabilities. Under large-scale RL training, these models exhibit emergent human-like thinking abilities characterized by deep, extended, and strategic reasoning. This allows them to explore diverse strategies, reflect on their decisions, revisit previous steps, and verify their conclusions. Such human-like thinking markedly improves accuracy, especially on complex reasoning tasks.

\paragraph{Efficient Thinking}
Given that o1-like models aim to mimic human thought processes, the efficiency of their reasoning is critical to their performance on challenging problems. Just as human thinking can occasionally be inefficient, models may face similar issues. For instance,~\citet{overthinking} study the problem of \textbf{overthinking} in o1-like LLMs, where models waste substantial computational resources revisiting trivial or self-evident paths, leading to inefficiency in {\em simple problems}. Conversely, our focus lies on the underexplored problem of \textbf{underthinking}, which occurs when a model fails to deeply explore promising paths, instead frequently switching strategies prematurely, resulting in computational waste. 
This inefficiency becomes especially pronounced when tackling {\em challenging problems}.
We assert that truly intelligent systems must learn to adaptively allocate their computational resources, concentrating on paths that are both promising and challenging.
During evaluations on the NPR Sunday Puzzle Challenge, \citet{anderson2025phd} note that o1-like LLMs often produce ``I give up'' mid-reasoning, prematurely ending their thought process and leading to incorrect outcomes. We view this as a form of underthinking.

To promote efficient reasoning, our subsequent work has also explored methods that limit unnecessary computation. Building on the observation that incorrect reasoning often leads to longer responses, Raoof and Dimakis propose Laconic decoding,\footnote{\url{https://x.com/AlexGDimakis/status/1885447830120362099}} which employs a shortest-of-n strategy to boost accuracy. \citet{muennighoff2025s1} control test-time compute by cutting off the model’s reasoning earlier or by repeatedly adding ``Wait'' to extend the reasoning when the model tries to end prematurely.
\citet{arora2025training} propose a modified reinforcement learning objective encouraging models to produce correct answers with relatively short CoT, thereby minimizing inference costs while preserving accuracy. Similarly, \citet{wu2025more} introduce Length-filtered Vote, which adaptively identifies the best CoT length for majority voting, excluding CoTs that are either overly short or unnecessarily long.

\subsection{Manipulating Decoding Penalties}
The role of penalty mechanisms in Natural Language Processing decoding has garnered significant attention. Traditional decoding methods, such as greedy search and beam search, focus primarily on maximizing the likelihood of generated sequences without considering the broader implications of the outputs. However, researchers have identified various shortcomings in these approaches, leading to the exploration of penalty mechanisms to enhance the quality of generated text.

Length normalization is a widely used strategy to adjust decoding penalties. \citet{jean2015montreal, koehn2017six, tu2017reconstruction, murray2018correcting} highlighted that length normalization and length penalties can prevent models from generating overly verbose or excessively brief translations, leading to improved fluency and adequacy.
In addition, \citet{tu2016modeling} introduced coverage penalties in neural machine translation to mitigate the problems of ``over-translation'' and ``under-translation'' by integrating a coverage metric that penalizes repeated attention to tokens. Along this direction, \citet{gnmt} proposed a coverage penalty in decoding to encourage the generation of an output that is most likely to cover all the words in the source sentence. \citet{see2017get} incorporated the concept of coverage into the summarization task by modeling the coverage content in summarization outputs.

In this paper, we adjust decoding penalties to address the problem of underthinking. Our approach encourages the model to maintain its original line of reasoning and engage in deeper thought processes, avoiding frequent shifts in strategy and superficial reasoning patterns. To the best of our knowledge, we are the first to investigate the effectiveness of decoding penalties in mitigating the underthinking issue.

\section{Conclusion}

In this work, we investigated underthinking in o1-like LLMs, identifying it as a significant factor limiting their performance on challenging reasoning tasks. Through comprehensive analysis, we observed that these models frequently abandon promising reasoning paths prematurely, leading to inefficient problem-solving and lower accuracy. We introduced a novel metric to quantify underthinking by assessing token efficiency in incorrect responses. To mitigate this issue, we proposed a decoding strategy with a {\color{ngreen} t}hought sw{\color{ngreen} i}tching {\color{ngreen} p}enalty (\textsc{Tip}), which encourages models to thoroughly explore each reasoning thought before considering alternatives. Our empirical results demonstrate that \textsc{Tip} effectively reduces underthinking and enhances accuracy across difficult mathematical and scientific problem sets without necessitating additional model training. 

This work contributes to a deeper understanding of reasoning processes in o1-like LLMs and provides a practical approach to align their problem-solving capabilities.
By addressing underthinking, we aim to bring models closer to human-like deep thinking, efficiently utilizing computational resources to achieve higher accuracy on complex tasks.
Future directions include exploring adaptive mechanisms within models to self-regulate thought transitions and further improving reasoning efficiency in o1-like LLMs.

\bibliography{ref}
\bibliographystyle{colm2024_conference}

\end{document}